\documentclass[runningheads,a4paper]{llncs}

\usepackage{amssymb}
\usepackage{graphicx}
\usepackage{amsfonts,amsmath}
\usepackage{float}
\usepackage{epstopdf}
\usepackage[latin1]{inputenc}
\usepackage{algorithm,algorithmic}
\usepackage{longtable}
\usepackage{url}

\begin{document}

\mainmatter  

\title{Double Ramp Loss Based Reject Option Classifier}


%
%
\author{Naresh Manwani\inst{1}%
\and Kalpit Desai\inst{2}\and Sanand Sasidharan\inst{1}\and Ramasubramanian Sundararajan\inst{3}}
%

\institute{Data Mining Lab, GE Global Research, JFWTC, Whitefield, Bangalore-560066,\\
(\email{Naresh.Manwani@ge.com, Sanand.Sasidharan@ge.com}),
\and
Bidgely, Bangalore (\email{kvdesai@gmail.com}),
\and
Sabre Airline Solutions, Bangalore (\email{gs.ramsu@gmail.com})}

%
%

\maketitle

\def \bw {\tilde{\mathbf{w}}}
\def \bx {\tilde{\mathbf{x}}}
\def \bz {\tilde{\mathbf{Z}}}
\def \bX {\tilde{\mathbf{X}}}
\def \xx {\mathbf{x}}
\def \ee {\mbox{\boldmath $1$}}
\def \ww {\mathbf{w}}
\def \alphaa  {\mbox{\boldmath $\alpha$}}
\def \xii  {\mbox{\boldmath $\xi$}}
\def \betaa  {\mbox{\boldmath $\beta$}}
\def \gammaa  {\mbox{\boldmath $\gamma$}}
\def \zz {\mbox{\boldmath $0$}}
\def \R {\mathbb{R}}
\def \U {\mathcal{U}}
\def \uu {\mathbf{u}}
\def \bb {\mathbf{b}}
\def \rr {\mathbf{r}}
\def \ll {\mathbf{l}}
\def \zzz {\mathbf{z}}

\begin{abstract}
We consider the problem of learning reject option classifiers. The goodness of a reject option classifier is quantified using $0-d-1$ loss function wherein a loss $d \in (0,.5)$ is assigned for rejection. In this paper, we propose {\em double ramp loss} function which gives a continuous upper bound for $(0-d-1)$ loss. Our approach is based on minimizing
regularized risk under the double ramp loss using {\em difference of convex (DC) programming}. We show the effectiveness of our approach through experiments on synthetic and benchmark datasets. Our approach performs better than the state of the art reject option classification approaches.
\end{abstract}

\section{Introduction}
\label{sec:intro}
The primary focus of classification problems has been on algorithms that return a prediction on every example. However, in many real life situations, it may be prudent to {\em reject} an example rather than run the risk of a costly potential mis-classification. Consider, for instance, a physician who has to return a diagnosis for a patient based on the observed symptoms and a preliminary examination. If the symptoms are either ambiguous, or rare enough to be unexplainable without further investigation, then the physician might choose not to risk misdiagnosing the patient (which might lead to further complications).
He might instead ask for further medical tests to be performed, or refer the case to an appropriate specialist.
Similarly, a banker, when faced with a loan application from a customer, may choose not to decide on the basis of the available information, and ask for a credit bureau score.
While the follow-up actions might vary (asking for more features to describe the example, or using a different classifier), the principal response in these cases is to ``reject'' the example. This paper focuses on the manner in which this principal response is decided, i.e., which examples should a classifier reject, and why? From a geometric standpoint, we can view the classifier as being possessed of a decision surface (which separates points of different classes) as well as a rejection surface.
The size of the rejection region impacts the proportion of cases that are likely to be rejected by the classifier, as well as the proportion of predicted cases that are likely to be correctly classified. A well-optimized classifier with a reject option is the one which minimizes the rejection rate as well as the mis-classification rate on the predicted examples.

Let $\xx \in \R^p$ is the feature vector and $y\in \{-1,+1\}$ is the class label. Let $\mathcal{D}(\xx,y)$ be the joint distribution of $\xx$ and $y$. A typical {\em reject option classifier} is defined using a bandwidth parameter ($\rho$) and a separating surface ($f(\xx)=0$). $\rho$ is the parameter which determines the rejection region. Then a reject option classifier $h(f(\xx),\rho)$ is formed as:
\begin{eqnarray}
\label{eq:compositeclassifier}
h(f(\xx),\rho)=\begin{cases}
1 & \text{if }  f(\xx)>   \rho\\
0 & \text{if } |f(\xx)|   \leq\rho\\
-1 & \text{if } f(\xx) < -\rho
\end{cases}
\end{eqnarray}
The reject option classifier can be viewed as two parallel surfaces with the rejection area in between. The goal is to determine $f(\xx)$ as well
as $\rho$ simultaneously.
The performance of this classifier is evaluated using $L_{0-d-1}$ \cite{Wegkamp2011,Radu2006} which is
\begin{eqnarray}\label{eq:disc-loss}
L_{0-d-1}(f(\xx),y,\rho)=
\begin{cases}
1, & \text{if }yf(\xx)<-\rho\\
d, & \text{if }|f(\xx)|\leq \rho \\
0, & \text{otherwise}
\end{cases}
\end{eqnarray}
In the above loss, $d$ is the cost of rejection.
If $d=0$, then we will always reject. When $d>.5$, then we will never reject (because expected loss of random labeling is 0.5).
Thus, we always take $d\in(0,.5)$.

To learn a reject option classifier, the expectation of $L_{0-d-1}(.,.,.)$ with respect to $\mathcal{D}(\xx,y)$ ({\em risk}) is minimized. Since $\mathcal{D}(\xx,y)$ is fixed but unknown, the empirical risk minimization principle is used.
The risk under $L_{0-d-1}$ is minimized by {\em generalized Bayes discriminant} \cite{Radu2006,Chow1970}, which is as below:
\begin{eqnarray}\label{eq:bayes}
f_d^*(\xx)=\begin{cases}
-1, &  \text{if }P(y=1|\xx) <d\\
0, &   \text{if }d \leq P(y=1|\xx) \leq 1-d\\
1, &   \text{if } P(y=1|\xx) > 1-d
\end{cases}
\end{eqnarray}
$h(f(\xx),\rho)$ (equation~(\ref{eq:compositeclassifier})) is shown to be infinite sample consistent with respect to the generalized Bayes classifier $f^*_d(\xx)$ described in equation (\ref{eq:bayes}) 
\cite{Yuan2010}.

\begin{table}
\vspace{-10pt}
\begin{center}
\begin{tabular}{|p{1.2in}|p{3.5in}|}
\hline 
{\bf Loss Function}   &  {\bf Definition}  \\ \hline
Generalized~Hinge  & $L_{\text{GH}}(f(\xx),y)=\begin{cases}
1-\frac{1-d}{d}yf(\xx), & \text{if }yf(\xx) <0\\
1-yf(\xx), & \text{if } 0\leq yf(\xx) <1\\
0, & \text{otherwise}
\end{cases}$      \\
\hline 
Double~Hinge   & $L_{\text{DH}}(f(\xx),y)=\max[-y(1-d)f(\xx)+H(d),-ydf(\xx)+H(d),0]$\\
&  where $H(d)=-d\log(d)-(1-d)\log(1-d)$ \\
\hline  
\end{tabular}
\caption{Convex surrogates for $L_{0-d-1}$.}\label{tab:losses}
\end{center}
\vspace{-30pt}
\end{table}

Since minimizing the risk under $L_{0-d-1}$ is computationally cumbersome, convex surrogates for $L_{0-d-1}$ have been proposed. {\em Generalized
hinge loss} $L_{\text{GH}}$ (see Table~\ref{tab:losses}) is a convex surrogate for $L_{0-d-1}$ \cite{Wegkamp2011,Wegkamp2007,Bartlett2008}. 
It is shown that a minimizer of risk under $L_{\text{GH}}$ is consistent to the generalized Bayes classifier
\cite{Bartlett2008}.
{\em Double hinge loss} $L_{\text{DH}}$ (see Table~\ref{tab:losses}) is another convex surrogate for $L_{0-d-1}$ \cite{Grandvalet2008}. Minimizer of the risk under $L_{\text{DH}}$ is shown to be {\em strongly universally consistent} to the generalized Bayes classifier \cite{Grandvalet2008}.

\begin{figure}[t]
\begin{center}
\begin{tabular}{cc}
\includegraphics[scale=0.32]{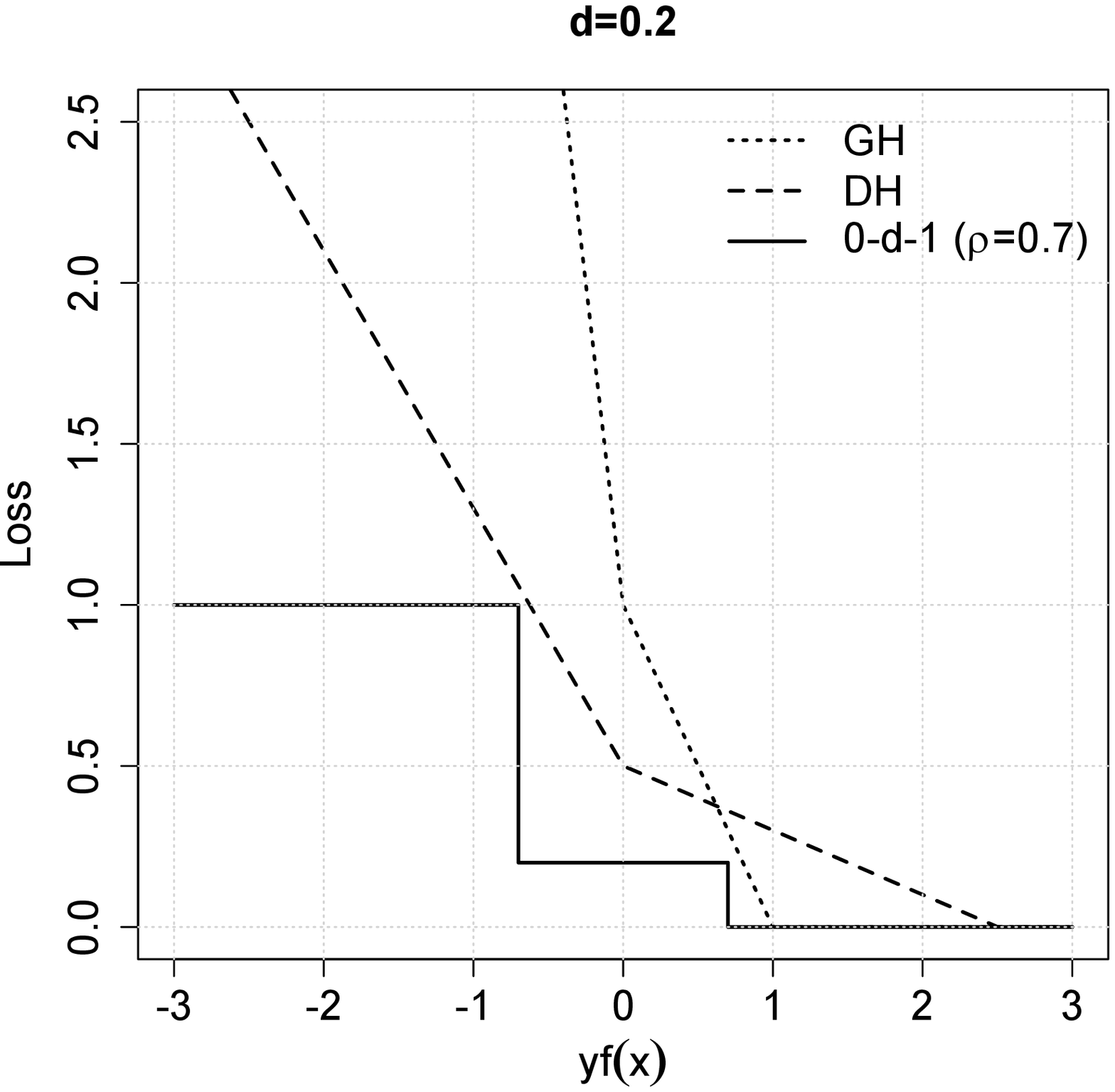} & \includegraphics[scale=0.32]{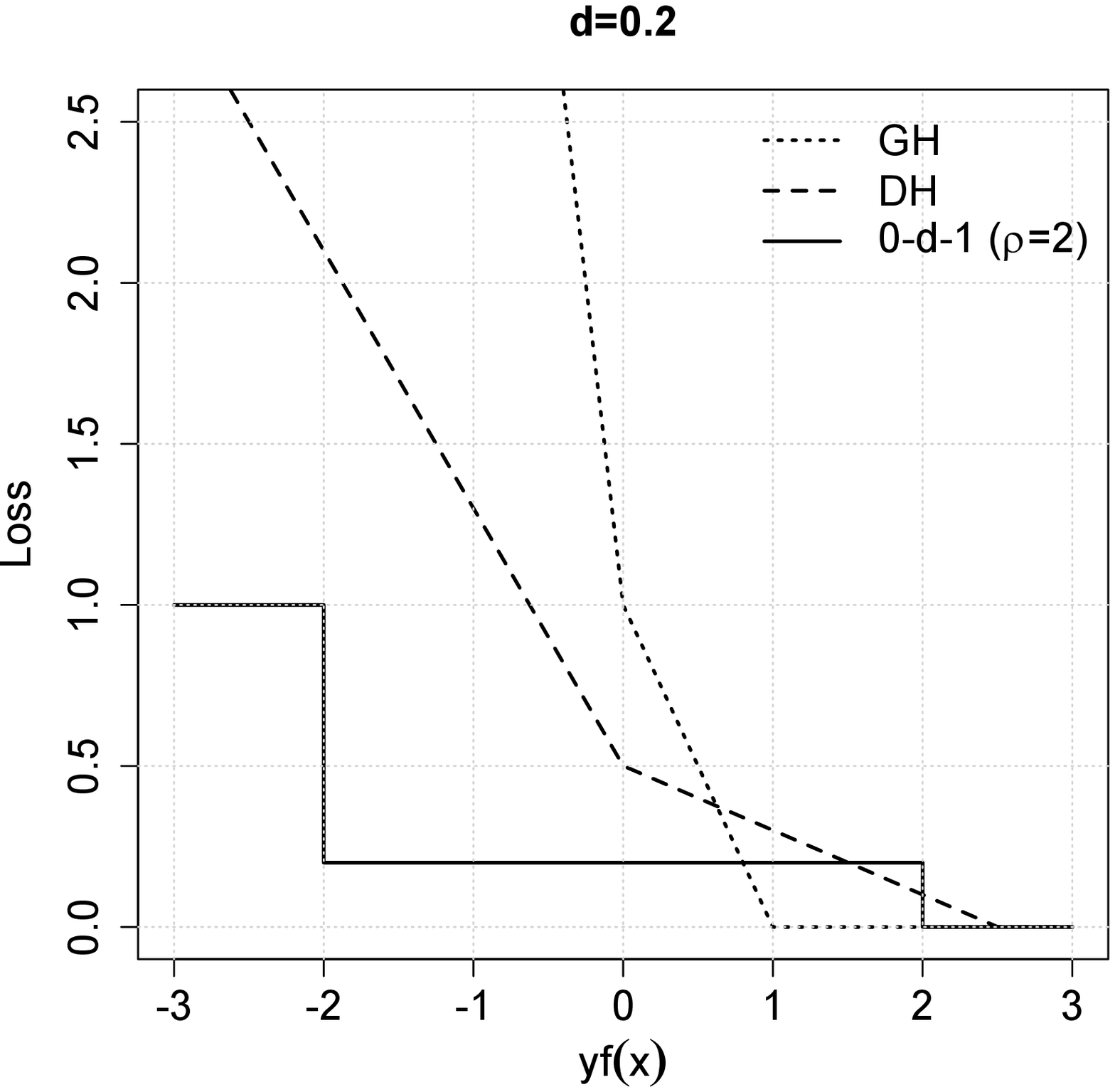}\\
(a) & (b)
\end{tabular}
\caption{\footnotesize{$L_{\text{GH}}$ and $L_{\text{DH}}$ for $d=0.2$. (a) For $\rho = 0.7$, both the losses upper bound the $L_{0-d-1}$. For $\rho =2$, both the losses fail to upper bound $L_{0-d-1}$. $L_{\text{GH}}$ and $L_{\text{DH}}$ both increase linearly even in the rejection region than being flat.}}\label{fig:losses} 
\end{center}
\vspace{-25pt}
\end{figure}

We observe that these convex loss functions have some limitations.
For example, $L_{\text{GH}}$ is a convex upper bound to $L_{0-d-1}$ provided $\rho < 1-d$ and $L_{\text{DH}}$ forms an upper bound to $L_{0-d-1}$ provided $\rho \in (\frac{1-H(d)}{1-d},\frac{H(d)-d}{d})$ (see Fig.~\ref{fig:losses}). 
Also, both $L_{\text{GH}}$ and $L_{\text{DH}}$ increase linearly in the rejection region
instead of remaining constant. These convex losses can become unbounded for misclassified examples with the scaling of parameters of $f$.
Moreover, limited experimental results are shown to validate the practical significance of these losses \cite{Wegkamp2011,Wegkamp2007,Bartlett2008,Grandvalet2008}.
A non-convex formulation for learning reject option classifier is proposed in \cite{fumera2002}. However, theoretical guarantees for the approach proposed in \cite{fumera2002} are not known. While learning a reject option classifier, one has to deal with the overlapping class regions as well as the presence of outliers.
SVM and other convex loss based approaches are less robust to label noise and outliers in the data \cite{Naresh2013}.
It is shown that ramp loss based risk minimization is more robust to noise \cite{GhoshMS14}.

Motivated from this, we propose {\em double ramp loss} $(L_{\text{DR}})$ which incorporates a different loss value for rejection.
$L_{\text{DR}}$ forms a continuous nonconvex upper bound for $L_{0-d-1}$ and overcomes many of the issues of convex surrogates of $L_{0-d-1}$. To learn a reject option classifier, we minimize the regularized risk under $L_{\text{DR}}$ which becomes an instance of difference of convex (DC) functions.
To minimize such a DC function, we use difference of convex programming approach \cite{Pham1997}, which essentially solves a sequence of convex programs.
The proposed method has following advantages over the existing approaches: (1) the proposed loss function $L_{\text{DR}}$ gives a tighter upper bound to the $L_{0-d-1}$,
(2) $L_{\text{DR}}$ requires no constraint on $\rho$ unlike $L_{\text{GH}}$ and $L_{\text{DH}}$,
(3) our approach can be easily kernelized for dealing with nonlinear problems.

The rest of the paper is organized as follows. In Section~\ref{sec:approach} we define the {\em double ramp loss} function $(L_{\text{DR}})$ and discuss its properties.
Then we discussed the proposed formulation based on risk minimization under $L_{\text{DR}}$. In Section~\ref{sec:SolutionMethodology} we derive the algorithm for learning reject option classifier based on regularized risk minimization under $(L_{\text{DR}})$ using DC programming. We present experimental results in Section~\ref{sec:Experiments}.
We conclude the paper with the discussion in Section~\ref{sec:conclusions}.

\section{Proposed Approach}\label{sec:approach}
Our approach for learning classifier with reject option is based on
minimizing regularized risk under $L_{\text{DR}}$ (double ramp loss).
\subsection{Double Ramp Loss}
\label{sec:DR}
We define double ramp loss function as a continuous upper bound for $L_{0-d-1}$.
This loss function is defined as a sum of two ramp loss functions as follows:
\begin{eqnarray}
\nonumber L_{\text{DR}}(f(\xx),y,\rho) &=&  
\frac{d}{\mu} \Big{[}\big{[}\mu-yf(\xx)+\rho\big{]}_+-\big{[}-\mu^2-yf(\xx)+\rho\big{]}_+\Big{]} \\
 && +\frac{(1-d)}{\mu}\Big{[}\big{[}\mu-yf(\xx)-\rho\big{]}_+- \big{[}-\mu^2-yf(\xx)-\rho\big{]}_+\Big{]}
\end{eqnarray}

\begin{figure}
\vspace{-20pt}
      \begin{center}
			\includegraphics[scale=0.35]{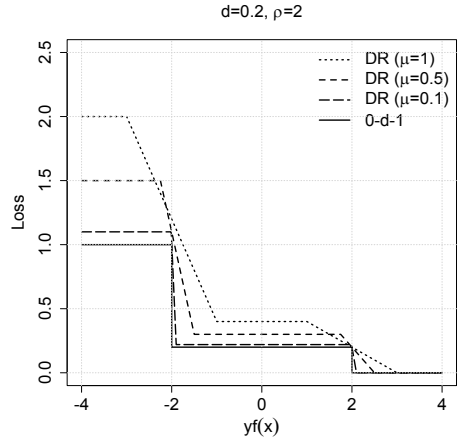} 
				\caption{$L_{\text{DR}}$ and $L_{0-d-1}$ : $\forall \mu \geq 0, \rho\geq 0$, $L_{\text{DR}}$
			is an upper bound for $L_{0-d-1}$.}
			\label{fig:DRL}
			\end{center}
						\vspace{-20pt}
\end{figure}
where $[a]_+=\max(0,a)$. $\mu \in (0,1]$ defines the slope of ramps in the loss function. $d\in (0,.5)$ is the cost of rejection and $\rho \geq 0$ is the parameter which defines the size of the rejection region around the classification boundary $f(\xx)=0$.\footnote{While $L_{\text{DR}}$ is parametrized by $\mu$ and $d$ as well, we omit them for the sake of notational consistency.} As in $L_{0-d-1}$, $L_{\text{DR}}$ also considers the region $[-\rho,\rho]$ as rejection region. Fig.~\ref{fig:DRL} shows $L_{\text{DR}}$ for $d=0.2, \rho=2$ with different values of $\mu$.
\begin{theorem}
\label{thm1}
(1) $L_{\text{DR}} \geq L_{0-d-1},\;\forall \mu>0,\rho \geq 0$.
(2) $\lim_{\mu \rightarrow 0} L_{\text{DR}}(f(\xx),\rho,y) = L_{0-d-1}(f(\xx),\rho,y)$.
(3) In the rejection region $yf(\xx) \in (\rho-\mu^2,-\rho+\mu)$, the loss remains constant, that is $L_{\text{DR}}(f(\xx),y,\rho)=d(1+\mu)$.
(4) For $\mu >0$, $L_{\text{DR}} \leq (1+\mu),\;\forall \rho \geq 0,\;\forall d \geq 0$.
(5) When $\rho =0$, $L_{\text{DR}}$ is same as $\mu$-ramp loss (\cite{Ong2012})used for classification problems without rejection option.
(6) $L_{\text{DR}}$ is a non-convex function of $(yf(\xx),\rho)$.
\end{theorem}
The proof of Theorem~\ref{thm1} is provided in Appendix~\ref{app:sec1}. We see that $L_{\text{DR}}$ does not put any restriction
on $\rho$ for it to be an upper bound of $L_{0-d-1}$.
Thus, $L_{\text{DR}}$ is a general ramp loss function which also allows rejection option.
\subsection{Risk Formulation Using $L_{\text{DR}}$}
\label{sec:formulation}
Let $\mathcal{S}=\{(\xx_n,y_n),\;n=1\ldots N\}$ be the training dataset, where $\xx_n \in \R^p,\;y_n\in\{-1,+1\},\;\forall n$.
As discussed, we minimize regularized risk under $L_{\text{DR}}$ to find a reject option classifier.
In this paper, we use $l_2$ regularization. Let $\Theta=[\ww^T\;\;\;b\;\;\;\rho]^T$. Thus, for $f(\xx)=(\ww^T\phi(\xx)+b)$, regularized risk under double ramp loss is
\begin{eqnarray}
\label{eq:Rwbrho1}
\nonumber R(\Theta) 
&=& \frac{1}{2}||\ww||^2 + C \sum_{n=1}^N L_{\text{DR}}(y_n,\ww^T\phi(\xx_n)+b)\\
\nonumber
 &= &\frac{1}{2}||\ww||^2 + \frac{C}{\mu} \sum_{n=1}^N \Big{\{}d\big{[}\mu-y_nf(\xx_n)+\rho\big{]}_+ -d\big{[}-\mu^2-y_nf(\xx_n)+\rho\big{]}_+  \\
\nonumber && + (1-d)\big{[}\mu-y_nf(\xx_n)-\rho\big{]}_+ -(1-d)\big{[}-\mu^2-y_nf(\xx_n)-\rho\big{]}_+\Big{\}}\\
\nonumber &=&\frac{1}{2}||\ww||^2 + \frac{C}{\mu} \sum_{n=1}^N \Big{\{}d\big{[}\mu-y_nf(\xx_n)+\rho\big{]}_+ +(1-d)\big{[}\mu-y_nf(\xx_n)-\rho\big{]}_+\\
\nonumber & &- d \big{[}-\mu^2-y_nf(\xx_n)+\rho\big{]}_+ -(1-d)\big{[}-\mu^2-y_nf(\xx_n)-\rho\big{]}_+\Big{\}}
\end{eqnarray}
where $C$ is regularization parameter. While minimizing $R(\Theta)$, no non-negativity condition on $\rho$ is required due to the following lemma.
\begin{lemma}\label{lemma1}
At the minimum of $R(\Theta)$, $\rho$ must be non-negative.
\end{lemma}
Prood of the above lemma is provided in Appendix~\ref{app:sec2}.

\section{Solution methodology}
\label{sec:SolutionMethodology}
$R(\Theta)$ (equation~(\ref{eq:Rwbrho1})) is a nonconvex function of $\Theta$. However, $R(\Theta)$ can be written as $R(\Theta)=R_1(\Theta)-R_2(\Theta)$,
where $R_1(\Theta)$ and $R_2(\Theta)$ are convex functions of $\Theta$. 
\begin{eqnarray}
\nonumber R_1(\Theta) &= & \frac{1}{2}||\ww||^2 + \frac{C}{\mu} \sum_{n=1}^N \Big{[}d\big{[}\mu-y_nf(\xx_n)+\rho\big{]}_++(1-d)\big{[}\mu-y_nf(\xx_n)-\rho\big{]}_+\Big{]}\\
\nonumber R_2(\Theta) &= &\frac{C}{\mu} \sum_{n=1}^N \Big{[} d \big{[}-\mu^2-y_nf(\xx_n)+\rho\big{]}_++(1-d)\big{[}-\mu^2-y_nf(\xx_n)-\rho\big{]}_+\Big{]}
\end{eqnarray}
In this case, DC programming
guarantees to find a local optima of $R(\Theta)$ \cite{Pham1997}.
In the simplified DC algorithm \cite{Pham1997}, an upper bound of $R(\Theta)$ is found using the convexity property of $R_2(\Theta)$ as follows.
\begin{eqnarray}
\label{eq:upbound}
R(\Theta) \leq   R_1(\Theta)- R_2(\Theta^{(l)}) -(\Theta-\Theta^{(l)})^T \nabla R_2(\Theta^{(l)})=: ub(\Theta,\Theta^{(l)})
\end{eqnarray}
where $\Theta^{(l)}$ is the parameter vector after $(l)^{th}$ iteration,
$\nabla R_2(\Theta^{(l)})$ is a sub-gradient of $R_2$ at $\Theta^{(l)}$.
$\Theta^{(l+1)}$ is found by
minimizing $ub(\Theta,\Theta^{(l)})$. Thus, $R(\Theta^{(l+1)}) \leq ub(\Theta^{(l+1)},\Theta^{(l)}) \leq ub(\Theta^{(l)},\Theta^{(l)})=R(\Theta^{(l)})$. Which means, in every iteration, the DC program reduces the value of $R(\Theta)$.
\subsection{Learning Reject Option Classifier Using DC Programming}
\label{sec:algo-dev}
In this section, we will derive a DC algorithm for minimizing $R(\Theta)$. We initialize with $\Theta=\Theta^{(0)}$.
For any $l \geq 0$, we find $ub(\Theta,\Theta^{(l)})$ as an upper bound for $R(\Theta)$ (see equation~(\ref{eq:upbound})) as follows:
\begin{equation}
\nonumber ub(\Theta,\Theta^{(l)})=R_1(\Theta)-R_2(\Theta^{(l)}) -(\Theta-\Theta^{(l)})^T \nabla R_2(\Theta^{(l)})
\end{equation}
Given $\Theta^{(l)}$, we find $\Theta^{(l+1)}$ by minimizing the upper bound $ub(\Theta,\Theta^{(l)})$. Thus,
\begin{eqnarray}
\label{eq:bound-min}
 \Theta^{(l+1)} \in \arg\min_{\Theta}\; ub(\Theta,\Theta^{(l)}) = \arg\min_{\Theta}\; R_1(\Theta) - \Theta^T\nabla R_2(\Theta^{(l)})
\end{eqnarray}
where $\nabla R_2(\Theta^{(l)})$ is the subgradient of $R_2(\Theta)$ at $\Theta^{(l)}$.
We choose $\nabla R_2(\Theta^{(l)})$ as:
\begin{align}
\nonumber \nabla R_2(\Theta^{(l)})=\sum_{n=1}^N \beta_n'^{(l)}[-y_n\phi(\xx_n)^T\;\;-y_n\;\;1]^T+\sum_{n=1}^N  \beta_n''^{(l)}[-y_n\phi(\xx_n)^T\;\;-y_n\;\;-1]^T
\end{align}
where
\begin{eqnarray}
\label{eq:sets}
\begin{cases}
\beta_n'^{(l)} = \frac{Cd}{\mu}\mathbb{I}_{\{y_n(\phi(\xx_n)^T\ww^{(l)} +b^{(l)})-\rho^{(l)} < -\mu^2\}} \\
\beta_n''^{(l)} = \frac{C(1-d)}{\mu}\mathbb{I}_{\{y_n(\phi(\xx_n)^T\ww^{(l)} +b^{(l)})+\rho^{(l)} < -\mu^2\}}
\end{cases} 
\end{eqnarray}
For $f(\xx)=(\ww^T\phi(\xx)+b$, we rewrite the upper bound minimization problem described in equation~(\ref{eq:bound-min}) as follows,
\begin{eqnarray}
\nonumber P^{(l+1)} &=  \min_{\Theta}  & R_1(\Theta) - \Theta^T\nabla R_2(\Theta^{(l)})\\
\nonumber &=\smash{\displaystyle \min_{\ww,b,\rho}}&  \frac{1}{2}||\ww||^2 + \frac{C}{\mu} \sum_{n=1}^N \Big{[}d\big{[}\mu-y_nf(\xx_n)+\rho\big{]}_+ + (1-d)\big{[}\mu-y_nf(\xx_n)-\rho\big{]}_+\Big{]} \\
\nonumber &&  +\sum_{n=1}^N \beta_n'^{(l)} [y_nf(\xx_n)-\rho] +\sum_{n=1}^N \beta_n''^{(l)} [y_nf(\xx_n)+\rho]
\end{eqnarray}
Note that $P^{(l+1)}$ is a convex optimization problem where the optimization variables are $(\ww,b,\rho)$.
We rewrite $P^{(l+1)}$ as
\begin{eqnarray}
\nonumber P^{(l+1)}=&\smash{\displaystyle \min_{\ww,b,\xii',\xii'',\rho}} & \frac{1}{2}||\ww||^2 + \frac{C}{\mu} \sum_{n=1}^N\big{[} d\xi_n'+(1-d)\xi_n''\big{]} + \sum_{n=1}^N \beta_n'^{(l)} [y_n(\ww^T\phi(\xx_n) + b)-\rho]\\
\nonumber &&  +\sum_{n=1}^N \beta_n''^{(l)} [y_n(\ww^T\phi(\xx_n)+b)+\rho]\\
\nonumber & s.t. &y_n(\ww^T\phi(\xx_n)+b)\geq \rho+\mu-\xi_n',\;\;\; \xi_n' \geq 0,\;\;\; n=1\ldots N\\
\nonumber & &y_n(\ww^T\phi(\xx_n)+b)\geq -\rho+\mu-\xi_n'',\;\;\; \xi_n''\geq 0 \;\;\; n=1\ldots N
\end{eqnarray}
where $\xii'=[\xi_1'\;\;\xi_2'\ldots \xi_N']^T$ and $\xii''=[\xi_1''\;\;\xi_2''\ldots \xi_N'']^T$.
The dual optimization problem $D^{(l+1)}$ of $P^{(l+1)}$ is as follows.
\begin{eqnarray}\label{eq:dual}
\nonumber D^{(l+1)} = & \smash{\displaystyle \min_{\gammaa',\gammaa''}}& \frac{1}{2}\sum_{n=1}^N\sum_{m=1}^Ny_ny_m(\gamma'_n+\gamma_n'')(\gamma'_m+\gamma_m'')k(\xx_n,\xx_m)
 -\mu\sum_{n=1}^N (\gamma_n' + \gamma_n'')\\
\nonumber &s.t. & \begin{cases}
-\beta_n'^{(l)} \leq \gamma_n' \leq \frac{Cd}{\mu} - \beta_n'^{(l)} & n=1\ldots N\\
-\beta_n''^{(l)} \leq \gamma_n'' \leq \frac{C(1-d)}{\mu}-\beta_n''^{(l)} & n=1\ldots N\\
 \sum_{n=1}^N y_n(\gamma_n'+\gamma_n'') = 0 \;\; \sum_{n=1}^N (\gamma_n' - \gamma_n'') = 0 &
\end{cases}
\end{eqnarray}
where $\gammaa'=[\gamma_1'\;\;\gamma_2' \ldots \ldots \gamma_n']^T$
and $\gammaa''=[\gamma_1''\;\;\gamma_2'' \ldots \ldots \gamma_n'']^T$ are dual variables.
The derivation of dual $D^{(l+1)}$ can be seen in Appendix~\ref{app:sec3}.
At the optimality of $P^{(l+1)}$, $\ww$ can be found as $\ww=\sum_{n=1}^Ny_n(\gamma_n' + \gamma_n'')\phi(\xx_n)$.

Since $P^{(l+1)}$ has quadratic objective and linear
constraints, it holds strong duality with $D^{(l+1)}$.
Solving $D^{(l+1)}$ is more useful as it can be easily kernelized for non-linear problems.
Behavior of $\gamma_n'$ and $\gamma_n''$ under different cases is as follows. 
\begin{eqnarray}
\nonumber 
\begin{cases}
y_n(\ww^T\phi(\xx_n) + b) - \mu > \rho   & \Rightarrow  \gamma_n' = -\beta_n'^{(l)};\;\;
\gamma_n'' = -\beta_n''^{(l)} \\
y_n(\ww^T\phi(\xx_n) + b) -\mu = \rho  & \Rightarrow  \gamma_n' \in \big{(}-\beta_n'^{(l)},\frac{Cd}{\mu}-\beta_n'^{(l)}\big{)};\;\; \gamma_n'' = -\beta_n''^{(l)} \\
y_n(\ww^T\phi(\xx_n) + b) -\mu \in (-\rho, \rho) & \Rightarrow  \gamma_n'  = \frac{Cd}{\mu}-\beta_n'^{(l)};\;\; \gamma_n'' = -\beta_n''^{(l)}  \\
y_n(\ww^T\phi(\xx_n) + b) - \mu = -\rho  & \Rightarrow  \gamma_n'  = \frac{Cd}{\mu}-\beta_n'^{(l)};\;\;
\gamma_n''  \in  \big{(}-\beta_n''^{(l)},\frac{C(1-d)}{\mu} - \beta_n''^{(l)} \big{)} \\
y_n(\ww^T\phi(\xx_n) + b) - \mu < -\rho & \Rightarrow   \gamma_n'  = \frac{Cd}{\mu}-\beta_n'^{(l)} 
;\;\; \gamma_n''  =  \frac{C(1-d)}{\mu} - \beta_n''^{(l)} 
\end{cases}
\end{eqnarray}

\subsection{Finding $b^{(l+1)}$ and $\rho^{(l+1)}$}\label{sec:thresh}
The dual optimization problem above gives dual variables $\gammaa'^{(l+1)}$ and $\gammaa''^{(l+1)}$ using which the normal vector is found as
$\ww^{(l+1)}=\sum_{n=1}^N (\gamma_n'^{(l+1)}+\gamma_n''^{(l+1)})y_n\phi(\xx_n)$.
To find $b^{(l+1)}$ and $\rho^{(l+1)}$, we consider
$\xx_n \in \text{SV}'^{(l+1)} \cup \text{SV}''^{(l+1)}$, where
\begin{eqnarray}
\nonumber \text{SV}'^{(l+1)} &=&  \{\xx_n\;|\; y_n(\phi(\xx_n)^T\ww^{(l+1)}+b^{(l+1)}) =  \rho^{(l+1)} + \mu\}\\
\nonumber \text{SV}''^{(l+1)} &=&  \{\xx_n\;|\; y_n(\phi(\xx_n)^T\ww^{(l+1)}+b^{(l+1)}) = -\rho^{(l+1)} + \mu\}
\end{eqnarray}
We already saw that
\begin{enumerate}
\item If $\xx_n \in \text{SV}'^{(l+1)}$, then $\gamma_n'^{(l+1)} \in \big{(}-\beta_n'^{(l)},\frac{Cd}{\mu}-\beta_n'{(l)}\big{)}$ and $\gamma_n''^{(l+1)}=-\beta_n''^{(l)}$
\item If $\xx_n \in \text{SV}''^{(l+1)}$, then $\gamma_n'^{(l+1)}=\frac{Cd}{\mu}-\beta_n'^{(l)}$ and $\gamma_n''^{(l+1)} \in \big{(}-\beta_n''^{(l)},\frac{C(1-d)}{\mu}-\beta_n''^{(l)}\big{)}$
\end{enumerate}
We solve the system of linear equations corresponding to sets $\text{SV}'^{(l+1)}$ and $\text{SV}''^{(l+1)}$ for identifying $b^{(l+1)}$ and $\rho^{(l+1)}$.

\subsection{Summary of the Algorithm}
We fix $d\in [0,.5]$, $\mu\in(0,1]$ and $C$ and initialize the parameter vector $\Theta$ as $\Theta^{(0)}$.
In any iteration $(l)$, we find $\beta_n'^{(l)},\beta_n''^{(l)},\;n=1\ldots N$ (see equation~(\ref{eq:sets}))using $\Theta^{(l)}$.
We use $\beta_n'^{(l)},\beta_n''^{(l)},\;n=1\ldots N$ and
solve $D^{(l+1)}$ to find $\gammaa'^{(l+1)},\gammaa''^{(l+1)}$. $\ww^{(l+1)}$ is found as $\ww^{(l+1)}=\sum_{n=1}^Ny_n(\gamma_n'^{(l+1)}+\gamma_n''^{(l+1)})\phi(\xx_n)$. We find $b^{(l+1)}$ and $\rho^{(l+1)}$ as described in Section~\ref{sec:thresh}.
Thus, we have found $\Theta^{(l+1)}$. Using $\Theta^{(l+1)}$, we now find $\beta_n'^{(l+1)},\beta_n''^{(l+1)},\;n=1\ldots N$.  
We repeat the above two steps until the parameter vector $\Theta$ changes significantly. More formal description of our algorithm is provided in Algorithm~\ref{algo2}.

\begin{algorithm}
\caption{Learning Reject Option Classifier by Minimizing $R(\Theta)$}
\label{algo2}
\begin{algorithmic}
\STATE {\bf Input : }$d\in[0,.5],\;\mu\in(0,1],\;C>0$, $\mathcal{S}$\;
\STATE {\bf Output : }$\ww^*,b^*,\rho^*$
\STATE {\bf Initialize} $\ww^{(0)},b^{(0)},\rho^{(0)}$, $l=0$\;
\REPEAT
\STATE \textbf{Compute} $\beta_n'^{(l)} = \frac{Cd}{\mu}\mathbb{I}_{\{y_n(\phi(\xx_n)^T\ww^{(l)} +b^{(l)})-\rho^{(l)} < -\mu^2\}}$ \\ $\beta_n''^{(l)} = \frac{C(1-d)}{\mu}\mathbb{I}_{\{y_n(\phi(\xx_n)^T\ww^{(l)} +b^{(l)})+\rho^{(l)} < -\mu^2\}}$
\STATE \textbf{Find} $\gammaa'^{(l+1)},\gammaa''^{(l+1)}$ by solving $D^{(l+1)}$ described in equation~(\ref{eq:dual})
\STATE \textbf{Find} $\ww^{(l+1)}=\sum_{n=1}^Ny_n(\gamma_n'^{(l+1)}+\gamma_n''^{(l+1)})\phi(\xx_n)$
\STATE \textbf{Find} $b^{(l+1)}$ and $\rho^{(l+1)}$
by solving the system of linear equations corresponding to sets $\text{SV}_1^{(l+1)}$ and $\text{SV}_2^{(l+1)}$, where
\begin{eqnarray}
\nonumber \text{SV}'^{(l+1)}  &=&  \{\xx_n\;|\; y_n(\phi(\xx_n)^T\ww^{(l+1)}+b^{(l+1)}) = \rho^{(l+1)} + \mu\}\\
\nonumber \text{SV}''^{(l+1)}  &=&  \{\xx_n\;|\; y_n(\phi(\xx_n)^T\ww^{(l+1)}+b^{(l+1)}) = -\rho^{(l+1)} + \mu\}
\end{eqnarray}
\UNTIL{convergence of $\Theta^{(l)}$}
\end{algorithmic}
\end{algorithm}

\subsection{$\gammaa'$ and $\gammaa''$ at the Convergence of Algorithm~\ref{algo2}}
\label{sec:properties}
\begin{figure}
\vspace{-20pt}
\begin{center}
\begin{tabular}{cc}
\includegraphics[scale=.32]{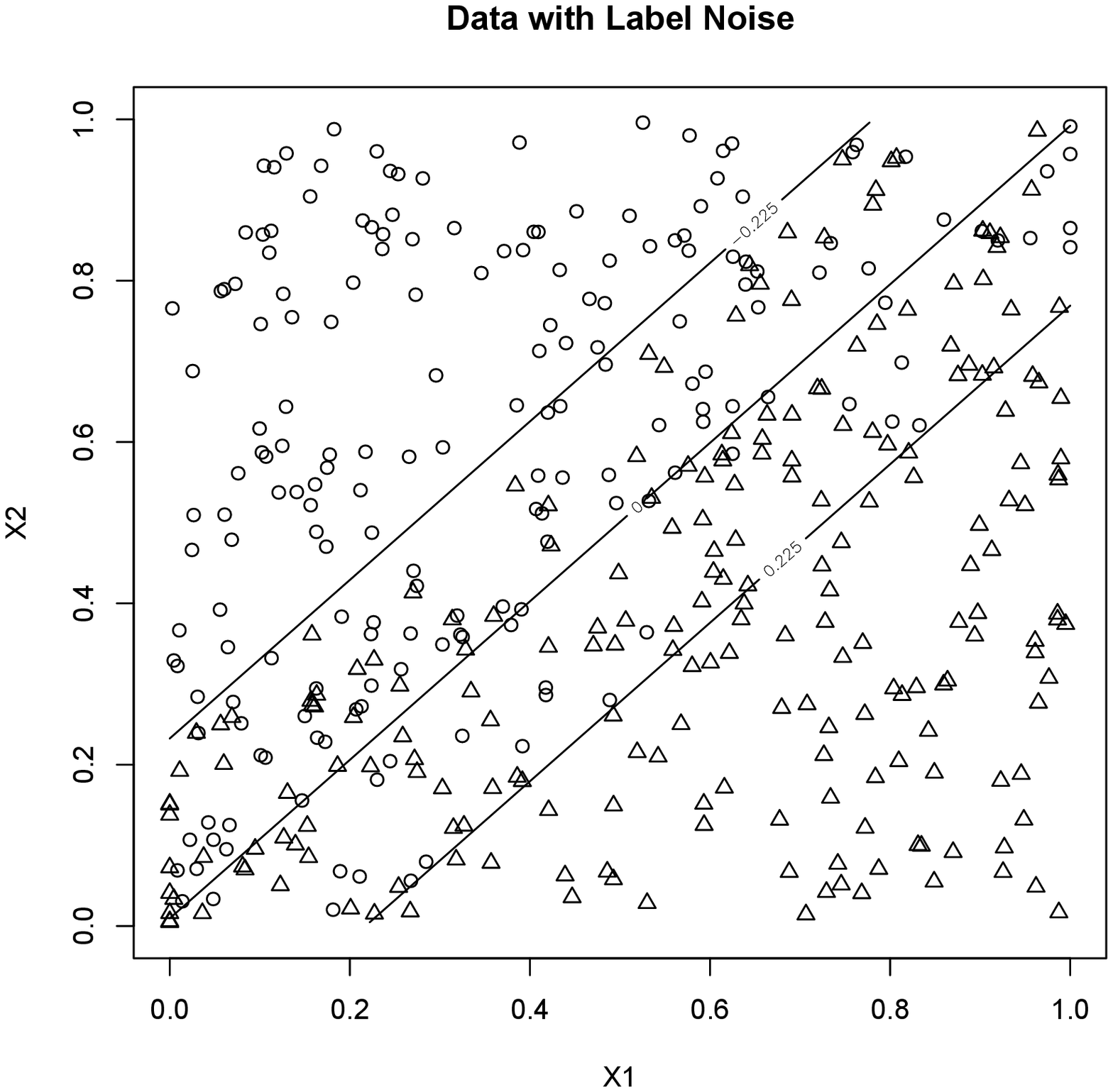}  & \includegraphics[scale=.32]{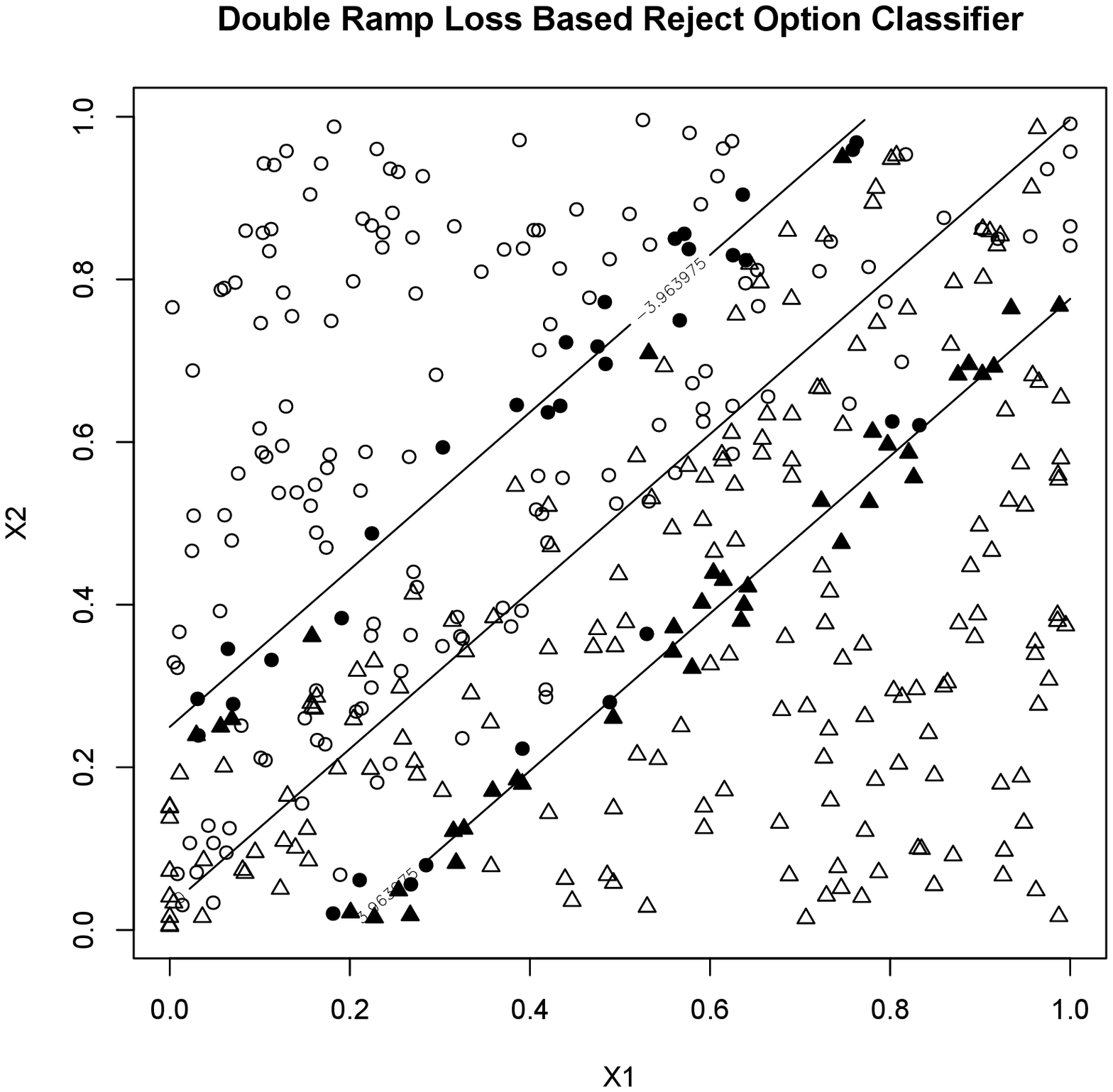}
\end{tabular}
\caption{\footnotesize{Figure on left shows that label noise affects points near the true classification boundary. Classes are represented using empty {\em circles} and {\em triangles}. Figure on right shows reject option classifier learnt using the proposed $L_{\text{DR}}$ based approach ($C=100$, $\mu=1$, $d=.2$). Filled {\em circles} and {\em triangles} represent the support vectors.}}\label{fig:ex1-syndata-coupled}
\end{center}
\vspace{-20pt}
\end{figure}

At the convergence of Algorithm~\ref{algo2}, let $\gamma_n'^*,\gamma_n''^*,\;n=1\ldots N$ become the values of the dual variables. The behavior of $\gamma_n'^*$ and $\gamma_n''^*$ is described in Table~\ref{table:behav_gamma}.
For any $\xx_n$, only one of $\gamma_n'^*$ and $\gamma_n''^*$ can be nonzero. We observe that parameters $\ww,b$ and $\rho$ are determined by the points
whose margin ($yf(\xx)$) is in the range $[\rho-\mu^2,\rho+\mu] \cup [-\rho-\mu^2,-\rho+\mu]$.
We call these points as {\em support vectors}.
We also see that $\xx_n$ for which $y_nf(\xx_n) \in (\rho+\mu,\infty)\cup (-\rho +\mu,\rho - \mu^2) \cup
(-\infty, -\rho - \mu^2)$, both $\gamma_n'^*,\gamma_n''^*=0$. Thus, points which are correctly classified with margin at least $(\rho+\mu)$, points falling close to the decision boundary with margin in the interval $(-\rho +\mu,\rho - \mu^2)$ and points misclassified with a high negative margin (less than $-\rho-\mu^2$), are ignored in the final classifier. Thus, our approach not only rejects points falling in the overlapping region of classes, it
also ignores potential outliers.
We illustrate these insights through experiments on a synthetic dataset as shown in Fig.~\ref{fig:ex1-syndata-coupled}. 400 points are uniformly sampled from the square region $[0\;\;1]\times [0\;\;1]$.
We consider the diagonal passing through the origin as the separating surface and assign labels $\{-1,+1\}$ to all the points using it.
We changed the labels of 80 points inside the band (width=0.225) around the separating surface.
\begin{table}[t]
\begin{center}
\begin{tabular}{|p{2.5in}|p{.5in}|p{.7in}|}
\hline
Condition  & $\gamma_n'^* \in$ & $\gamma_n''^* \in$ \\
\hline
$y_n(\ww^T\phi(\xx_n)+b) \in (\rho+\mu,\infty)$       & 0                      & 0\\
\hline
$y_n(\ww^T\phi(\xx_n)+b) =  \rho + \mu$                 & $(0,\frac{Cd}{\mu})$   & 0\\
\hline
$y_n(\ww^T\phi(\xx_n)+b) \in [\rho-\mu^2,\rho+\mu)$   & $\frac{Cd}{\mu}$       & 0\\
\hline
$y_n(\ww^T\phi(\xx_n)+b) \in (-\rho + \mu,\rho- \mu^2)$  & 0   & 0\\
\hline
$y_n(\ww^T\phi(\xx_n)+b)  = -\rho+\mu$         & 0         & $(0,\frac{C(1-d)}{\mu})$\\
\hline
$y_n(\ww^T\phi(\xx_n)+b)  \in [-\rho-\mu^2,-\rho+\mu)$     & 0     & $\frac{C(1-d)}{\mu}$\\
\hline
$y_n(\ww^T\phi(\xx_n)+b)  \in (-\infty,-\rho-\mu^2)$      & 0        & 0\\
\hline
\end{tabular}
\caption{Behavior of $\gammaa'^*$ and $\gammaa''^*$}\label{table:behav_gamma}
\end{center}
\vspace{-10pt}
\end{table}
Fig.~\ref{fig:ex1-syndata-coupled} shows the reject option classifier learnt using the proposed method. We see that the proposed approach learns the rejection region accurately. We also observe that all of the support vectors are near the two parallel hyperplanes.
\vspace{-10pt}
\section{Experimental Results}
\label{sec:Experiments}
\vspace{-10pt}
We show the effectiveness of our approach by showing its performance on several datasets. We also compare our approach with the approach proposed in \cite{Grandvalet2008}. 
\subsection{Dataset Description}
We report experimental results on 1 synthetic datasets and 2 datasets taken from UCI ML repository \cite{Bache2013}.
\begin{enumerate}
\item {\bf Synthetic Dataset 1 : }Let $f_1$ and $f_2$ be two mixture density functions in $\R^2$ defined as follows:
\begin{eqnarray}
\nonumber f_1(\xx) = 0.45 \U([ 1, 0] \times [ 1, 1]) + 0.5 \U([ 4, 3] \times [0,1])
+ 0.05 \U([ 10, 0] \times [ 5, 5])\\
\nonumber f_2(\xx) = 0.45 \U([0, 1] \times [ 1, 1]) + 0.5 \U([9, 10] \times [ 1, 0])
+ 0.05 \U([0, 10] \times [ 5, 5])
\end{eqnarray}
where $\U(A)$ denotes the uniform density function with support set $A$.
We sample 150 points independently each from $f_1$
and $f_2$. We label
these points using the hyperplane with $\ww = [1\;\;\;0]^T$ and $b = 0$.
We choose 10\% of these points uniformly at random and flip their labels.
\item {\bf Synthetic Dataset 2 \cite{Hastie2001} :} $\mathbf{m}_{k1},k=1,\ldots,10$ were drawn from $\mathcal{N}((1,0)^T,I)$ and labeled as class $C_1$. Similarly, $\mathbf{m}_{k2},\;k=1,\ldots,10$ were drawn from $\mathcal{N}((0,1)^T,I)$ and labeled as class $C_2$. For each class, 100 observations were drawn from the following mixture distributions:
\[ f(\xx|C_i) = \sum_{k=1}^{10} \frac{1}{10} \mathcal{N}(\mathbf{m}_{ki},I/5),\;\;\;i=1,2\]
\item {\bf Ionosphere Dataset \cite{Bache2013} : }This dataset describes the problem of discriminating {\em good versus bad radars} based on whether they send some useful information about the Ionosphere. There are 34 variables and 351 observations.
\item {\bf Parkinsons Disease Dataset \cite{Bache2013} : }This dataset is used to discriminate people with Parkinsons disease from the healthy people. There are 22 features which are comprised of a range of biomedical voice measurements from individuals. There are 195 such feature vectors.
\end{enumerate} 

\subsection{Experimental Setup}
In the proposed $L_{\textbf{DR}}$ based approach, for solving the dual $D^{(l)}$ at every iteration, we have used the {\em kernlab} package \cite{kernlab} in {\bf R}.
We thank the authors of $L_{\text{DH}}$ based method \cite{Grandvalet2008} for providing the codes for their approach. 
For nonlinear problems, we use RBF kernel.
In our approach, we set $\mu=1$. $C$ and $\sigma$ (width parameter for RBF kernel)
are chosen using 10-fold cross validation. 

\subsection{Simulation Results}
For every dataset, we report results for values of $d$ in the interval $[0.05\;\;\;.5]$ with the step size of 0.05. For every value of $d$, we find the cross validation
risk (under $L_{0-d-1}$), \% accuracy on the non-rejected examples (Acc) and \% rejection rate (RR).
The results provided are based on 10 repetitions of 10-fold
cross validation (CV). We show the average values and standard
deviation (computed over the 10 repetitions).

\begin{figure}
\begin{center}
\begin{tabular}{ccc}
\includegraphics[scale=.23]{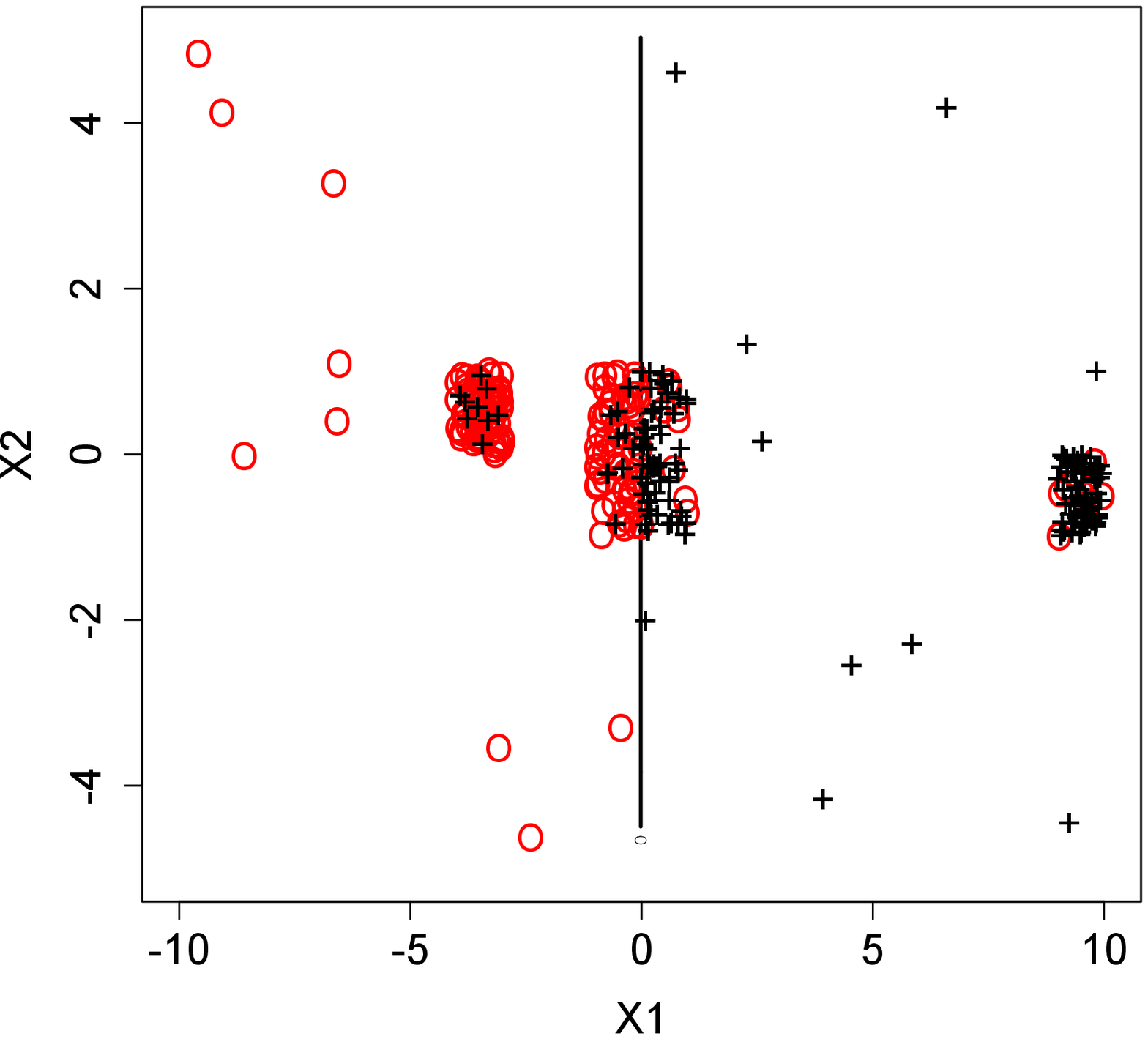}  & \includegraphics[scale=.23]{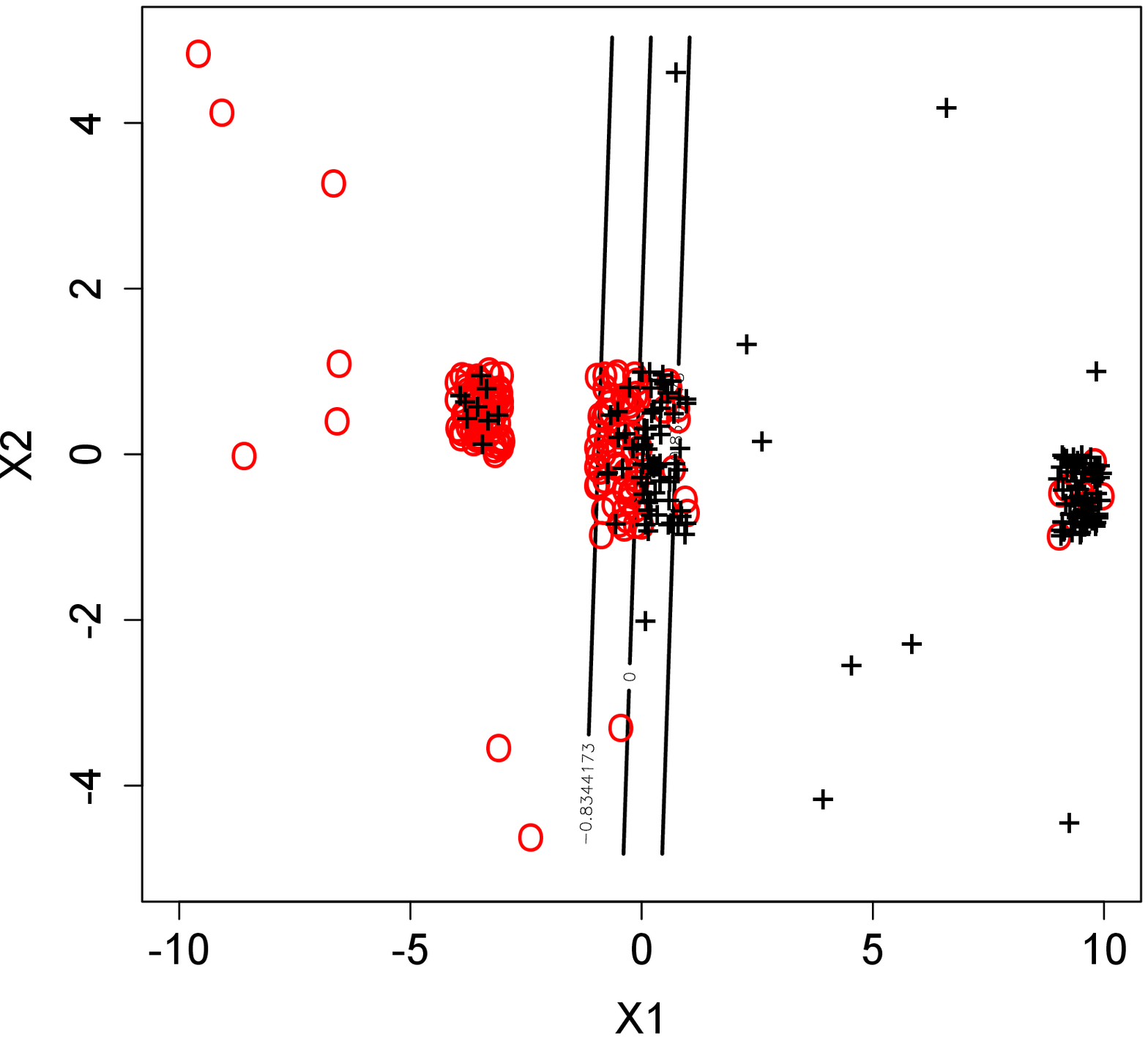}
& \includegraphics[scale=.23]{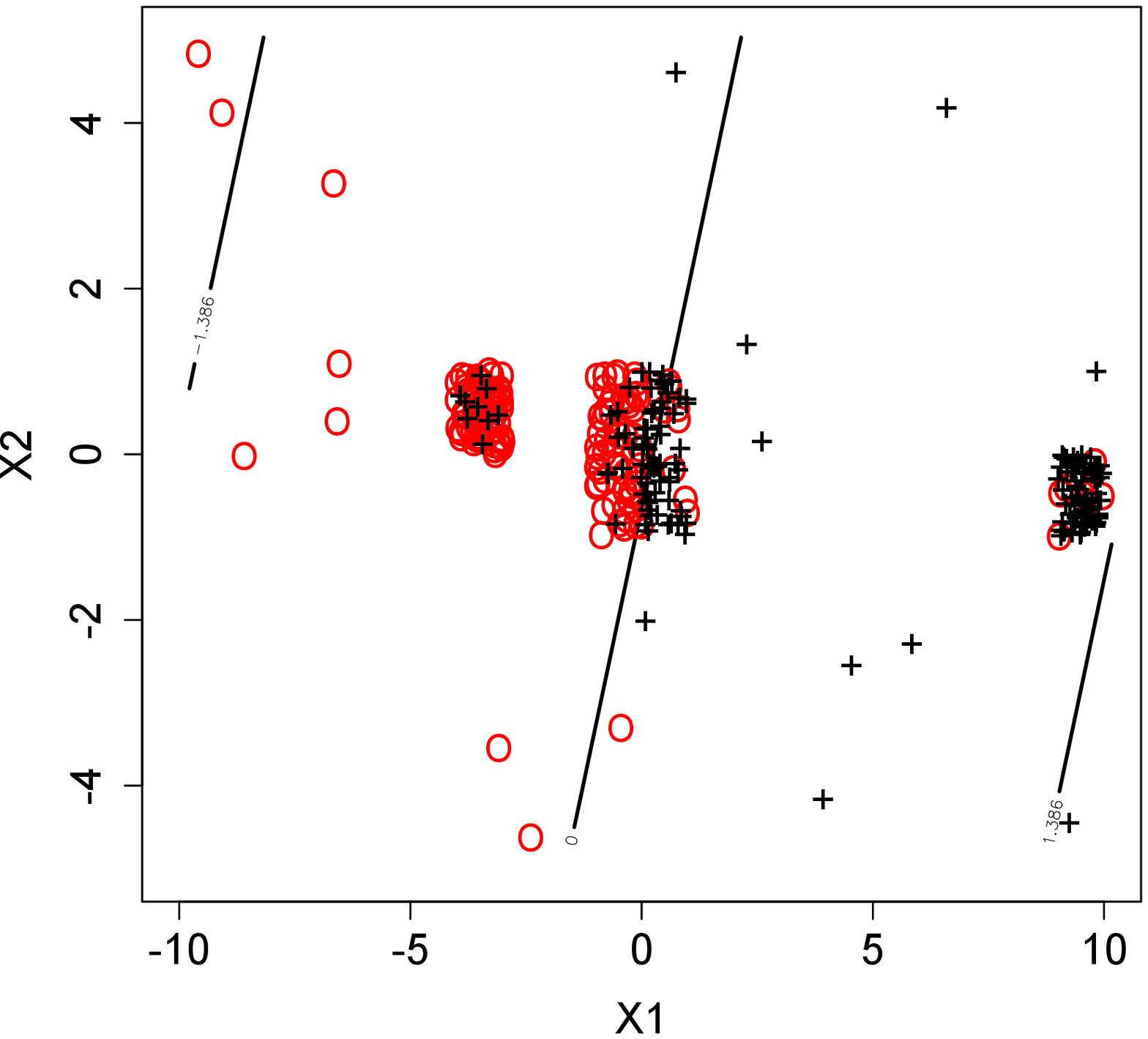}\\
(a) & (b)  & (c)
\end{tabular}
\caption{(a) Synthetic Dataset and the true classification boundary. Reject option classifiers learnt using (b) proposed $L_{DR}$ based approach for $d=0.2$, (c) $L_{DH}$ based approach for $d=0.2$.}\label{results_syndata1} 
\end{center}
\vspace{-10pt}
\end{figure}

We now discuss the experimental results.
Fig.~\ref{results_syndata1}(a) shows the Synthetic dataset and the true classification boundary. This dataset has some mislabeled points creating noise around the classification surface. Fig.~\ref{results_syndata1}(b) and (c) show the classifiers learnt using $L_{\text{DR}}$ and $L_{\text{DH}}$ based approaches respectively for $d=0.2$. We see that $L_{\text{DR}}$ based approach accurately finds the true classification boundary as oppose to $L_{\text{DH}}$ based approach. Also, the reject region found by $L_{\text{DR}}$
 based approach is covers the most ambiguous region unlike $L_{\text{DH}}$ based approach which rejects almost all the points.

Table~\ref{tab:Synth1}-\ref{tab:Park} show the experimental results on all the datasets. We observe the following:
\begin{enumerate}
\item We see that the proposed $L_{\text{DR}}$ based method outperforms $L_{\text{DH}}$ based approach in terms of the risk (expectation of $L_{0-d-1}$). For Synthetic dataset 1, except for $d=0.05$ and $0.1$, $L_{\text{DR}}$ based method has lower CV risk. For Synthetic dataset 2, both the approaches perform comparable to each other. For Ionosphere dataset, except for $d=0.2,0.25$ and $0.3$, $L_{\text{DR}}$ based method has lower CV risk. For Parkinsons dataset, $L_{\text{DR}}$ based method has lower CV risk except for $d=0.35$. 
\item We also observe that $L_{\text{DR}}$ based method outputs classifiers with significantly lesser rejection rate for all the datasets and for all values of $d$.
\end{enumerate}
Thus, for most of the cases, the proposed $L_{\text{DR}}$ based approach outputs classifiers with lesser risk. Moreover, the learnt classifier has always lesser rejection rate compared to the $L_{\text{DH}}$ based approach.

\begin{table}[t]
  \centering  
    \begin{tabular}{||p{.3in}||p{.75in}|p{.75in}|p{.7in}||p{.75in}|p{.7in}|p{.7in}||}
    		\hline
        \textbf{d}  &  \multicolumn{3}{c||}{\textbf{$L_{\text{DR}}$ ($C=2$)}}  & \multicolumn{3}{c||}{\textbf{$L_{\text{DH}}$ ($C=32$)}} \\
\hline
		    &  \textbf{Risk} & \textbf{RR} & \textbf{Acc on unrejected} &  \textbf{Risk} & \textbf{RR} & \textbf{Acc on unrejected}  \\
				\hline  
0.05 &  0.068$\pm$0.015 & 90.87$\pm$5.79    & 75.87$\pm$7.95  &   {\bf 0.05}             & 100              & NA\\
0.1	 &  0.138$\pm$0.023 & 70.35$\pm$12.18  & 79.05$\pm$6.87  &  {\bf 0.105}$\pm$0.002	 & 95.53$\pm$1.69  & 77.20$\pm$6.06\\
0.15 &  {\bf 0.135}$\pm$0.003 & 65.41$\pm$5.06   & 89.66$\pm$0.90  &  0.136             & 72.77$\pm$0.23 & 90.56$\pm$0.66\\
0.2  &  {\bf 0.155}$\pm$0.006	& 43.18$\pm$4.31   & 88.56$\pm$0.75   &  0.17              & 72.67           & 90.36$\pm$1.44\\
0.25 &  {\bf 0.164}$\pm$0.014 & 32.13$\pm$8.43  & 87.97$\pm$1.42   &  0.204$\pm$0.003   & 66.5$\pm$1.7   & 91$\pm$0.74\\
0.3  &  {\bf 0.148}$\pm$0.012 & 13.23$\pm$7.52  & 87.67$\pm$0.69  &  0.197             & 46.73$\pm$0.14 & 89.37$\pm$0.32\\
0.35 &  {\bf 0.134}$\pm$0.005 & 4.57$\pm$1.80    & 87.68$\pm$0.23   &  0.21$\pm$0.002    & 43.33$\pm$0.65 & 90.02$\pm$0.38\\
0.4  &  {\bf 0.131}$\pm$0.003 & 1.51$\pm$0.56   & 87.29$\pm$0.30   &  0.21$\pm$0.006    & 31.17$\pm$1.26 & 87.41$\pm$0.55\\
0.45 &  {\bf 0.128}$\pm$0.002 & 0.86$\pm$0.45    & 87.45$\pm$0.25  &  0.265$\pm$0.008   & 9.13$\pm$1.1  & 75.58$\pm$0.98\\
0.5  &  {\bf 0.136}$\pm$0.01 & 0                  & 86.41$\pm$0.99   &  0.297$\pm$0.004   & 0                & 70.27$\pm$0.44 \\
    \hline 
    \end{tabular}%
		\caption{Comparison results on Synthetic Dataset 1 (linear classifiers for both the approaches).}
  \label{tab:Synth1}%
\end{table}%

\begin{table}[t]
  \centering  
    \begin{tabular}{||p{.3in}||p{.75in}|p{.75in}|p{.7in}||p{.75in}|p{.7in}|p{.7in}||}
    		\hline
        \textbf{d}  &  \multicolumn{3}{c||}{\textbf{$L_{\text{DR}}$ ($C=64,\;\gamma = 0.25$)}}  & \multicolumn{3}{c||}{\textbf{$L_{\text{DH}}$ ($C=64,\;\gamma = 0.25$)}} \\
\hline
		    &  \textbf{Risk} & \textbf{RR} & \textbf{Acc on unrejected} &  \textbf{Risk} & \textbf{RR} & \textbf{Acc on unrejected}  \\
				\hline  
0.05 &	0.046$\pm$0.006 &  79.5$\pm$1.47   &  97.56$\pm$2.92  &  0.046$\pm$0.004  &  86.5$\pm$0.82  & 97.26$\pm$3.8 \\
0.1  &  {\bf 0.096}$\pm$0.006 &  75.45$\pm$1.12  &  92.80$\pm$2.35  &  0.1$\pm$0.005    &  76.35$\pm$1.13 & 91.65$\pm$2.0 \\
0.15 &  0.15$\pm$0.012  &  64.3$\pm$2.32   &  86.40$\pm$2.35  &  {\bf 0.139}$\pm$0.01   &  52.3$\pm$2.02  & 87.6$\pm$2.4 \\
0.2  &  0.182$\pm$0.01  &  51.2$\pm$1.90   &  84.79$\pm$1.99  &  {\bf 0.162}$\pm$0.007	&  40.35$\pm$1.68 & 86.75$\pm$1.22 \\
0.25 &  0.193$\pm$0.008 &  30.3$\pm$1.01   &  83.56$\pm$1.33  &  {\bf 0.18}$\pm$0.008   &  31.25$\pm$1.65 & 85.74$\pm$1.47 \\
0.3  &  0.190$\pm$0.005 &  16.4$\pm$1.74   &  83.47$\pm$0.75  &  {\bf 0.183}$\pm$0.013  &  18.35$\pm$2.85	& 84.4$\pm$1.2 \\
0.35 &  0.178$\pm$0.006 &  6.85$\pm$1.43   &  83.49$\pm$0.69  &  0.178$\pm$0.008  &  10.65$\pm$1.42 & 84.21$\pm$0.80 \\
0.4  &  {\bf 0.171}$\pm$0.012 &  2.6$\pm$1.26    &  83.51$\pm$1.2   &  0.177$\pm$0.006  &  5.75$\pm$0.68  & 83.75$\pm$0.76 \\
0.45 &  {\bf 0.168}$\pm$0.011 &  0.65$\pm$0.41   &  83.42$\pm$1.06  &  0.182$\pm$0.008  &  2.95$\pm$0.9   & 82.61$\pm$0.87  \\
0.5  &  {\bf 0.178}$\pm$0.014 & 	0	  &  82.2$\pm$1.36  & 0.184$\pm$0.009 & 	0	  & 81.65$\pm$0.88\\
    \hline 
   \end{tabular}%
	\caption{Comparison Results on Synthetic Dataset 2 (nonlinear classifiers using RBF kernel for both the approaches).}
  \label{tab:Synth2}%
\end{table}%

\begin{table}[t]
  \centering  
    \begin{tabular}{||p{.3in}||p{.75in}|p{.75in}|p{.7in}||p{.75in}|p{.7in}|p{.7in}||}
    		\hline
        \textbf{d}  &  \multicolumn{3}{c||}{\textbf{$L_{\text{DR}}$ ($C=2,\;\gamma = 0.125$)}}  & \multicolumn{3}{c||}{\textbf{$L_{\text{DH}}$ ($C=16,\;\gamma = 0.125$)}} \\
\hline
		    &  \textbf{Risk} & \textbf{RR} & \textbf{Acc on unrejected} &  \textbf{Risk} & \textbf{RR} & \textbf{Acc on unrejected}  \\
				\hline  
0.05  &{\bf 0.025}$\pm$0.002  &  34.84$\pm$0.92  &  98.94$\pm$0.31  &  0.029      &   52.61$\pm$0.73  &  99.47$\pm$0.06 \\
0.1   &{\bf 0.027}$\pm$0.003  &  8.81$\pm$0.32   &  97.99$\pm$0.33  &  0.047$\pm$0.002  &  43.44$\pm$0.85  &  99.46$\pm$0.17 \\
0.15  &{\bf 0.039}$\pm$0.003  &  5.78$\pm$0.57   &  96.81$\pm$0.29  &  0.042$\pm$0.003  &  24.02$\pm$1.62  &  99.3$\pm$0.37 \\
0.2   &0.044$\pm$0.001  &  3.46$\pm$0.51   &  96.18$\pm$0.15  &  {\bf 0.04}$\pm$0.002    & 17.43$\pm$0.59  &  99.42$\pm$0.25 \\
0.25  &0.047$\pm$0.002  &  1.76$\pm$0.41   &  95.68$\pm$0.23  &  {\bf 0.046}$\pm$0.001  &  14.47$\pm$0.79  &  98.9$\pm$0.16 \\
0.3   &0.052$\pm$0.003  &  0.92$\pm$0.46   &  95.08$\pm$0.35  &  {\bf 0.051}$\pm$0.003  &  12.57$\pm$0.75  &  98.56$\pm$0.31 \\
0.35  &{\bf 0.051}$\pm$0.003  &  0.03$\pm$0.09   &  94.88$\pm$0.29  &  0.054$\pm$0.002	 &  9.33$\pm$0.59   &  97.72$\pm$0.21 \\
0.4   &{\bf 0.051}$\pm$0.002  &  0               &  94.95$\pm$0.24  &  0.054$\pm$0.003  &  6.72$\pm$0.86   &  97.09$\pm$0.35 \\
0.45  &{\bf 0.054}$\pm$0.002  &  0	              &  94.64$\pm$0.21  &  0.055$\pm$0.003  &  3.53$\pm$0.41   &  95.97$\pm$0.36 \\
0.5   &{\bf 0.054}$\pm$0.001  &  0               &  94.62$\pm$0.13  &  0.055$\pm$0.005  &  0                & 94.55$\pm$0.47 \\
    \hline 
    \end{tabular}%
		\caption{Comparison results on Ionosphere dataset (nonlinear classifiers using RBF kernel for both the approaches).}
  \label{tab:Ionosphere}%
\end{table}%

\begin{table}
  \centering  
    \begin{tabular}{||p{.3in}||p{.75in}|p{.75in}|p{.7in}||p{.75in}|p{.7in}|p{.7in}||}
    		\hline
        \textbf{d}  &  \multicolumn{3}{c||}{\textbf{$L_{\text{DR}}$ ($C=32$)}}  & \multicolumn{3}{c||}{\textbf{$L_{\text{DH}}$ ($C=32$)}} \\
\hline
		    &  \textbf{Risk} & \textbf{RR} & \textbf{Acc on unrejected} &  \textbf{Risk} & \textbf{RR} & \textbf{Acc on unrejected}  \\
				\hline  
0.05 & {\bf 0.031}$\pm$0.002  &  43.88$\pm$0.80  &  98.33$\pm$0.49  &  0.043$\pm$0.001  &  86.38$\pm$0.92  & 100	\\
0.1  & {\bf 0.051}$\pm$0.004  &  41.79$\pm$0.77  &  98.07$\pm$1.03  &  0.061$\pm$0.002  &  53.76$\pm$1.64  & 98.61$\pm$0.62 \\
0.15 & {\bf 0.071}$\pm$0.002  &  40.08$\pm$1.21  &  98.14$\pm$0.48  &  0.086$\pm$0.004  &  39.56$\pm$1.13  & 95.8$\pm$0.72 \\
0.2  & {\bf 0.095}$\pm$0.004  &  37.67$\pm$1.04  &  96.99$\pm$0.55  &  0.125$\pm$0.008  &  29.78$\pm$2.06  &  90.86$\pm$1.5 \\
0.25 & {\bf 0.133}$\pm$0.009  &  20.46$\pm$2.79  &  90.26$\pm$1.30  &  0.142$\pm$0.004  &  22.3$\pm$1.95  &  89.02$\pm$0.73 \\
0.3  & {\bf 0.129}$\pm$0.01   &  4.06$\pm$2.06   &  87.83$\pm$1.15  &  0.131$\pm$0.009  &  14.19$\pm$1.05 &  89.76$\pm$1.01 \\
0.35 & 0.134$\pm$0.007  &  2.49$\pm$1.04  &  87.19$\pm$0.76  &  {\bf 0.133}$\pm$0.004  &  9.97$\pm$1.18  &  89.10$\pm$0.57 \\
0.4 & {\bf 0.131}$\pm$0.008  &  0.56$\pm$0.44  &  87.06$\pm$0.75  &  0.133$\pm$0.006	&  6.10$\pm$1.62  & 88.53$\pm$0.92 \\
0.45 &  {\bf 0.133}$\pm$0.013  &  0.05$\pm$0.17  &  86.72$\pm$1.28  &  0.14$\pm$0.009 & 	2.92$\pm$1.09 & 86.96$\pm$1.05\\
0.5 & {\bf 0.133}$\pm$0.009  &  0  &  86.65$\pm$0.94  &  0.139$\pm$0.008 &  0	  &  86.06$\pm$0.76  \\
    \hline 
    \end{tabular}%
		\caption{Comparison results on Parkinsons Disease dataset (linear classifiers for both the approaches).}
  \label{tab:Park}%
\end{table}%

\section{Conclusion and Future Work}
\label{sec:conclusions}
In this paper, we have proposed a new loss function $L_{\text{DR}}$ ({\bf double ramp loss}) for learning the reject option classifier. $L_{\text{DR}}$ gives tighter upper bound for $L_{0-d-1}$ compared to
convex losses $L_{\text{DH}}$ and $L_{\text{GH}}$. 
Our approach learns the classifier by minimizing the regularized {\em risk} under the double ramp loss function which becomes an instance of
DC optimization problem. Our approach can also learn nonlinear classifiers by using appropriate kernel function.
Experimentally we have shown that our approach works superior to $L_{\text{DH}}$ based approach for learning reject option classifier.

\bibliography{mybiblio}
\bibliographystyle{plain}

\begin{appendix}
\section{Proof of Theorem~\ref{thm1}}
\label{app:sec1}
\begin{eqnarray}
\nonumber L_{\text{DR}}(f(\xx),\rho,y) &=&  
\frac{d}{\mu} \Big{[}\big{[}\mu-yf(\xx)+\rho\big{]}_+-\big{[}-\mu^2-yf(\xx)+\rho\big{]}_+\Big{]}\\
\nonumber && +\frac{(1-d)}{\mu}\Big{[}\big{[}\mu-yf(\xx)-\rho\big{]}_+- \big{[}-\mu^2-yf(\xx)-\rho\big{]}_+\Big{]}
\end{eqnarray}
\begin{enumerate}
\item Table~\ref{table:table1} shows that $L_{\text{DR}} \geq L_{0-d-1},\;\forall \mu>0,\rho \geq 0$.
\begin{table}[h]
\begin{center}
\begin{tabular}{|l|c|r|}
\hline
Interval  & $L_{\text{DR}}$ & $L_{0-d-1}$\\
\hline
$yf(\xx) \in [\rho+\mu,\infty)$                                          & 0                                    &  0                        \\
\hline
$yf(\xx) \in (\rho,\rho+\mu)$                                            & $\in (0,d)$                          &  0                        \\
\hline
$yf(\xx) \in (\rho-\mu^2,\rho]$                                          & $\in [d,(1+\mu)d)$                   &  $d$                      \\
\hline
$yf(\xx) \in [-\rho+\mu,\rho-\mu^2]$                                     & $(1+\mu)d$                           &  $d$                      \\
\hline
$yf(\xx) \in [-\rho,-\rho+\mu)$                                          & $\in ((1+\mu)d,(1+\mu)d+(1-d)]$      &  $d$                      \\
\hline
$yf(\xx) \in (-\rho-\mu^2,-\rho)$                                        & $\in ((1+\mu)d+(1-d),(1+\mu))$       & 1                         \\
\hline
$yf(\xx) \in (-\infty,-\rho-\mu^2]$                                      & $1+\mu$                              & 1                         \\
\hline
\end{tabular}
\caption{Proof for Theorem 1.(1).}\label{table:table1}
\end{center}
\end{table}
\item We need to show that $\lim_{\mu \rightarrow 0} L_{\text{DR}}(f(\xx),\rho,y) = L_{0-d-1}(f(\xx),\rho,y)$.
We first see the values that $L_{\text{DR}}$ take for different values of $yf(\xx)$. Table~\ref{table:table2} shows how $L_{\text{DR}}$ changes as a function of $yf(\xx)$.  
\begin{table}[h]
\begin{center}
\begin{tabular}{|l|r|}
\hline
Interval  & $L_{\text{DR}}$ \\
\hline
$yf(\xx) \in (\rho+\mu,\infty)$                                          & 0                                                        \\
\hline
$yf(\xx) \in [\rho-\mu^2,\rho+\mu]$                                      & $\frac{d}{\mu}(\mu-yf(\xx)+\rho)$                        \\
\hline
$yf(\xx) \in (-\rho+\mu,\rho-\mu^2)$                                     & $(1+\mu)d$                                               \\
\hline
$yf(\xx) \in [-\rho-\mu^2,-\rho+\mu]$                                    & $(1+\mu)d+\frac{(1-d)}{\mu}(\mu-yf(\xx)-\rho)$          \\
\hline
$yf(\xx) \in (-\infty,-\rho-\mu^2)$                                      & $1+\mu$                                                 \\
\hline
\end{tabular}
\caption{$L_{\text{DR}}$ in different intervals (Proof for Theorem~1.(iii))}\label{table:table2}
\end{center}
\end{table}

Now we take the limit $\mu \rightarrow 0$, which is shown in Table~\ref{table:table3}. We see that $\lim_{\mu \rightarrow 0}L_{\text{DR}} = L_{0-d-1}$.
\begin{table}[h]
\begin{center}
\begin{tabular}{|c|c|c|}
\hline
Interval  & $\lim_{\mu \rightarrow 0}L_{\text{DR}}$  & $L_{0-d-1}$\\
\hline
$yf(\xx) \in (\rho,\infty)$                                          & 0& 0                                                        \\
\hline
$yf(\xx) = \rho$                                                     &     $d$ & $d$                       \\
\hline
$yf(\xx) \in (-\rho,\rho)$                                           & $d$& $d$                                               \\
\hline
$yf(\xx) = -\rho$                                    & $1$   & $1$        \\
\hline
$yf(\xx) \in (-\infty,-\rho)$                                      & $1$& $1$                                                  \\
\hline
\end{tabular}
\caption{$\lim_{\mu \rightarrow 0}L_{\text{DR}}$ in different intervals (Proof for Theorem~1.(iii))}\label{table:table3}
\end{center}
\end{table}
\item In the rejection region $yf(\xx) \in (\rho-\mu^2,-\rho+\mu)$, the loss remains constant, that is $L_{\text{DR}}(f(\xx),\rho,y)=d(1+\mu)$. This can be seen in Table~\ref{table:table2}.
\item For $\mu >0$, $L_{\text{DR}} \leq (1+\mu),\;\forall \rho \geq 0,\;\forall d \geq 0$. This can be seen in Table~\ref{table:table2}.
\item When $\rho =0$, $L_{\text{DR}}$ becomes
\begin{eqnarray}
\nonumber L_{\text{DR}}(f(\xx),0,y) &=&\frac{d}{\mu} \Big{[}\big{[}\mu-yf(\xx)\big{]}_+-\big{[}-\mu^2-yf(\xx)\big{]}_+\Big{]}+\frac{(1-d)}{\mu}\Big{[}\big{[}\mu-yf(\xx)-\big{]}_+\\
\nonumber && - \big{[}-\mu^2-yf(\xx)\big{]}_+\Big{]} \\
\nonumber &=& \frac{1}{\mu} \Big{[}\big{[}\mu-yf(\xx)\big{]}_+-\big{[}-\mu^2-yf(\xx)\big{]}_+\Big{]}
\end{eqnarray}
which is same as the $\mu$-ramp loss function used for classification problems without rejection option.
\item We have to show that $L_{\text{DR}}$ is non-convex function of $(yf(\xx),\rho)$. From (iv), we know that $L_{\text{DR}}\leq (1+\mu)$. That is, $L_{\text{DR}}$ is bounded above. We show non-convexity of $L_{\text{DR}}$ by contradiction.

Let $L_{\text{DR}}$ be convex function of $(yf(\xx),\rho)$. Let $\zzz=(yf(\xx),\rho)$. We also rewrite $L_{\text{DR}}(f(\xx),\rho,y)$ as $L_{\text{DR}}(\zzz)$.
We choose two points $\zzz_1,\zzz_2$ such that $L_{\text{DR}}(\zzz_1)>L_{\text{DR}}(\zzz_2)$. Thus, from the definition of convexity, we have
\begin{eqnarray}
\nonumber L_{\text{DR}}(\zzz_1) \leq \lambda L_{\text{DR}}(\frac{\zzz_1-(1-\lambda)\zzz_2}{\lambda})+(1-\lambda)L_{\text{DR}}(\zzz_2)\;\;\;\forall\lambda\in(0,1)
\end{eqnarray}
Hence,
\begin{equation}
 \frac{L_{\text{DR}}(\zzz_1)-(1-\lambda)L_{\text{DR}}(\zzz_2)}{\lambda}\leq L_{\text{DR}}(\frac{\zzz_1-(1-\lambda)\zzz_2}{\lambda}) \nonumber
\end{equation}
Now, since $L_{\text{DR}}(\zzz_1)>L_{\text{DR}}(\zzz_2)$, 
\begin{equation}
\frac{L_{\text{DR}}(\zzz_1)-(1-\lambda)L_{\text{DR}}(\zzz_2)}{\lambda}=\frac{L_{\text{DR}}(\zzz_1)-L_{\text{DR}}(\zzz_2)}{\lambda}+L_{\text{DR}}(\zzz_2)\rightarrow \infty \;\;\;as \;\;\; \lambda\rightarrow0^+ \nonumber
\end{equation}
Thus $\lim_{\lambda \rightarrow 0^+} L_{\text{DR}}(\frac{\zzz_1-(1-\lambda)\zzz_2}{\lambda}) = \infty$.
But $L_{\text{DR}}$ is upper bounded by $(1+\mu)d$. This contradicts that $L_{\text{DR}}$ is convex.
\end{enumerate}

\section{Proof of Lemma~\ref{lemma1}}
\label{app:sec2}
Let $\Theta'=(\ww',b',\rho')$ minimizes $R(\Theta)$, where $\rho'<0$. Thus $-\rho'>0$.
Consider $\Theta''=(\ww',b',-\rho')$ as another point. 
\begin{eqnarray}
\nonumber R(\Theta') - R(\Theta'') & = & \frac{C(1-2d)}{\mu} \sum_{n=1}^N \Big{\{}-\big{[}\mu-y_nf(\xx_n)+\rho'\big{]}_+ + \big{[}-\mu^2-y_nf(\xx_n)+\rho'\big{]}_+ \\
\nonumber &&    + \big{[}\mu-y_nf(\xx_n)-\rho'\big{]}_+  -\big{[}-\mu^2-y_nf(\xx_n)-\rho'\big{]}_+\Big{\}}\\
\nonumber &=& C(1-2d) \sum_{n=1}^N \Big{\{}L_{ramp}(y_nf(\xx_n)+\rho')-L_{ramp}(y_nf(\xx_n)-\rho') \Big{\}}
\end{eqnarray}
where $L_{ramp}(t) = \frac{1}{\mu} ([\mu-t]_+ - [-\mu^2 - t]_+)$ is a monotonically non-increasing function of $t$ \cite{Ong2012}. Since $\rho'<0$, thus, $y_nf(\xx_n)+\rho' < y_nf(\xx_n)-\rho',\;\forall n$. This implies $L_{ramp}(y_nf(\xx_n)+\rho') \geq L_{ramp}(y_nf(\xx_n)-\rho'),\;\forall n$. Also $(1-2d) \geq 0$, since $0 \leq d \leq 0.5$. Thus $R(\Theta') - R(\Theta'') \geq 0$,
which contradicts that $\Theta'$ minimizes $R(\Theta)$.
Thus, at the minimum of $R(\Theta)$, $\rho$ must be non-negative.

\section{Derivation of Dual Optimization Problem $\mathcal{D}^{(l+1)}$}\label{app:sec3}
\begin{eqnarray}
\nonumber \mathcal{P}^{(l+1)} :  &\smash{\displaystyle \min_{\ww,b,\xii',\xii'',\rho}} & \frac{1}{2}||\ww||^2 + \frac{C}{\mu} \sum_{n=1}^N\big{[} d\xi_n'+(1-d)\xi_n''\big{]} + \sum_{n=1}^N \beta_n'^{(l)} [y_n(\ww^T\phi(\xx_n) + b)-\rho]\\
\nonumber &&  +\sum_{n=1}^N \beta_n''^{(l)} [y_n(\ww^T\phi(\xx_n)+b)+\rho]\\
\nonumber & s.t. &y_n(\ww^T\phi(\xx_n)+b)\geq \rho+\mu-\xi_n',\;\;\; \xi_n' \geq 0,\;\;\; n=1\ldots N\\
\nonumber & &y_n(\ww^T\phi(\xx_n)+b)\geq -\rho+\mu-\xi_n'',\;\;\; \xi_n''\geq 0 \;\;\; n=1\ldots N
\end{eqnarray}
The Lagrangian for above problem will be:
\begin{eqnarray}
\nonumber \mathcal{L}&=&\frac{1}{2}||\ww||^2 + \frac{C}{\mu} \sum_{n=1}^N\big{[} d\xi_n'+(1-d)\xi_n''\big{]}+
\sum_{n=1}^N \beta_n'^{(l)} [y_n(\ww^T\phi(\xx_n) + b)-\rho]+\\
\nonumber & & \sum_{n=1}^N \beta_n''^{(l)} [y_n(\ww^T\phi(\xx_n)+b)+\rho] + \sum_{n=1}^N\alpha_n'[\rho + \mu-\xi_n'-y_n(\ww^T\phi(\xx_n)+b)] -\sum_{n=1}^N\eta_n'\xi_n' \\
\nonumber & & +\sum_{n=1}^N\alpha_n''[-\rho + \mu-\xi_n''-y_n(\ww^T\phi(\xx_n)+b)] -\sum_{n=1}^N\eta_n''\xi_n''\end{eqnarray}
where $\alpha_n'$ is dual variable corresponding to constraint $y_n(\ww^T\phi(\xx_n)+b)\geq \rho+\mu-\xi_n'$, $\alpha''_n$ is dual variable
corresponding to $ y_n(\ww^T\phi(\xx_n)+b)\geq -\rho+\mu-\xi_n'$, $\eta_n'$ is dual variable corresponding to $\xi_n'\geq 0$, $\eta_n''$ is dual variable corresponding to
$\xi_n'' \geq 0$.
We take the gradient of Lagrangian with respect to the primal variables. By equating the gradient to zero,
we get the KKT conditions of optimality for this optimization problem.
\begin{eqnarray}
\nonumber \begin{cases}
\ww = \sum_{n=1}^N y_n[\alpha_n' + \alpha_n''-\beta_n'^{(l)}-\beta_n''{(l)}]\phi(\xx_n) &\\
\sum_{n=1}^N y_n[\alpha_n' + \alpha_n''-\beta_n'^{(l)}-\beta_n''{(l)}] & \\
\eta_n' + \alpha_n' =\frac{Cd}{\mu} & n=1\ldots N\\
\eta_n'' + \alpha_n'' =\frac{C(1-d)}{\mu} & n=1\ldots N\\
\sum_{n=1}^N[\alpha_n' - \alpha_n''  - \beta_n'^{(l)} + \beta_n''{(l)}] =0 &\\
\eta_n'\xi_n' = 0,\;\;\eta_n'\geq 0  & n=1\ldots N\\
\eta_n''\xi_n''=0,\;\;\eta_n''\geq 0  & n=1\ldots N\\
\alpha_n'[\mu-\xi'_n-y_n(\ww^T\phi(\xx_n)+b)+\rho]=0,\;\;\alpha_n'\geq 0  & n=1\ldots N\\
\alpha_n''[\mu-\xi''_n-y_n(\ww^T\phi(\xx_n)+b)-\rho]=0,\;\;\alpha''_n\geq 0  &  n=1\ldots N
\end{cases}
\end{eqnarray}
We make the dual optimization problem simpler by changing the variables in following way:
\begin{eqnarray}
\begin{cases}
\nonumber\gamma_n'= \alpha_n' - \beta_n'^{(l)},\;\;\; n=1\ldots N\\
\nonumber\gamma_n''=\alpha_n'' - \beta_n''^{(l)},\;\;\; n =1\ldots N
\end{cases}
\end{eqnarray}
By changing these variables, the new KKT conditions in terms of $\gammaa'$ and $\gammaa''$ are
\begin{eqnarray}
\nonumber \begin{cases}
\ww = \sum_{n=1}^N y_n(\gamma_n' + \gamma_n'')\phi(\xx_n)  &\\
\sum_{n=1}^N y_n(\gamma_n' + \gamma_n'')=0 & \\
\eta_n' + \gamma_n' + \beta_n'^{(l)} = \frac{Cd}{\mu} & n = 1\ldots N\\
\eta_n'' + \gamma_n'' + \beta_n''^{(l)} = \frac{C(1-d)}{\mu} & n = 1\ldots N \\
\sum_{n=1}^N(\gamma_n' - \gamma_n'') =0 &\\
\eta_n'\xi_n' = 0,\;\;\eta_n'\geq 0  &  n=1\ldots N\\
\eta_n''\xi_n''=0,\;\;\eta_n''\geq 0  & n=1\ldots N\\
(\gamma_n' + \beta_n'^{(l)})[\mu-\xi'_n-y_n(\ww^T\phi(\xx_n)+b)+\rho]=0,\;\;\gamma_n' + \beta_n'^{(l)}\geq 0  & n=1\ldots N\\
(\gamma_n'' + \beta_n''^{(l)})[\mu-\xi''_n-y_n(\ww^T\phi(\xx_n)+b)+\rho]=0,\;\;\gamma_n'' + \beta_n''^{(l)}\geq 0  & n=1\ldots N
\end{cases}
\end{eqnarray}
Using the KKT conditions in the Langarangian, we replace the primal variables $(\ww,b,\rho,\xii',\xii'')$ in terms of
the dual variables $(\gammaa',\gammaa'')$. The dual optimization problem $\mathcal{D}^{(l+1)}$ will become:
\begin{eqnarray}
\nonumber \mathcal{D}^{(l+1)} = & \smash{\displaystyle \min_{\gammaa',\gammaa''}}& \frac{1}{2}\sum_{n=1}^N\sum_{m=1}^Ny_ny_m(\gamma'_n+\gamma_n'')(\gamma'_m+\gamma_m'')k(\xx_n,\xx_m)
 -\mu\sum_{n=1}^N (\gamma_n'+ \gamma_n'')\\
\nonumber &s.t. & \begin{cases}
-\beta_n'^{(l)} \leq \gamma_n' \leq \frac{Cd}{\mu} - \beta_n'^{(l)} & n=1\ldots N\\
-\beta_n''^{(l)} \leq \gamma_n'' \leq \frac{C(1-d)}{\mu}-\beta_n''^{(l)} & n=1\ldots N\\
 \sum_{n=1}^N y_n(\gamma_n'+\gamma_n'') = 0 & \\
\sum_{n=1}^N (\gamma_n' - \gamma_n'')= 0 &
\end{cases}
\end{eqnarray}
where $\gammaa'=[\gamma_1'\;\;\gamma_2' \ldots \ldots \gamma_n']^T$
and $\gammaa''=[\gamma_1''\;\;\gamma_2'' \ldots \ldots \gamma_n'']^T$.
\end{appendix}
\end{document}